%% file: main.tex
\documentclass[conference]{IEEEtran}

\usepackage[utf8]{inputenc}
\usepackage{cite}
\usepackage{amsmath,amssymb,amsfonts}
\usepackage{algorithmic}
\usepackage{graphicx}
\usepackage{textcomp}
\usepackage{xcolor}
\usepackage{booktabs}
\usepackage[normalem]{ulem}
\useunder{\uline}{\ul}{}
\usepackage[capitalize]{cleveref}
\usepackage[textsize=tiny, textwidth=1.0cm, colorinlistoftodos]{todonotes}

\title{Designing Decision Support Systems for Emergency Response: Challenges and Opportunities}

\author{
	\IEEEauthorblockN{Geoffrey Pettet\IEEEauthorrefmark{1}, Hunter Baxter\IEEEauthorrefmark{1}, Sayyed Mohsen Vazirizade\IEEEauthorrefmark{1}, \\ Hemant Purohit\IEEEauthorrefmark{3}, Meiyi Ma\IEEEauthorrefmark{1}, Ayan Mukhopadhyay\IEEEauthorrefmark{1}, Abhishek Dubey\IEEEauthorrefmark{1}\\}
	\IEEEauthorblockA{
		\IEEEauthorrefmark{1}Department of Computer Science, Vanderbilt University, Nashville, TN 37235, USA 	\\ \IEEEauthorrefmark{3}Information Sciences and Technology Department, George Mason University, Fairfax, VA 22030, USA \\
		}
}

\date{February 2022}

\begin{document}

\maketitle

\input{Abstract}

\input{Introduction}

\input{DataCollection}
\input{IncidentForecasting}
\input{IncidentDetection}
\input{ResourceAllocation}
\input{Conclusion}

\bibliographystyle{IEEEtran}
\bibliography{refs.bib}

\end{document}

%% file: Abstract.tex
\begin{abstract}
Designing effective emergency response management (ERM) systems to respond to incidents such as road accidents is a major problem faced by communities. In addition to responding to frequent incidents each day (about 240 million emergency medical services calls and over 5 million road accidents in the US each year), these systems also support response during natural hazards. Recently, there has been a consistent interest in building decision support and optimization tools that can help emergency responders provide more efficient and effective response. This includes a number of principled subsystems that implement early incident detection, incident likelihood forecasting and strategic resource allocation and dispatch policies. In this paper, we highlight the key challenges and provide an overview of the approach developed by our team in collaboration with our community partners.
\end{abstract}

%% file: Introduction.tex
\section{Introduction} \label{sec:intro}
Designing effective emergency response management (ERM) systems to oversee incidents such as road accidents and fires is a challenge faced by communities across the globe. These systems must efficiently manage resources such as ambulances to respond quickly to incidents so that human and financial losses are minimized~\cite{jaldell2017important, jaldell2014time}. Over the last several decades, much attention has been given to studying emergency incidents and response, but emergency incidents still cause thousands of deaths and injuries and result in losses worth billions of dollars each year~\cite{crimeUS}. 

A fundamental shift in how these systems is operated has begun in the last decade. Critically, the use of deep learning methods and statistical methods that provide principled uncertainty-aware decision support systems are now being applied to this domain~\cite{mukhopadhyay2020review,MukhopadhyayDissertation2019}. These methods typically include five major components: (1) data curation components that clean and merge environmental features like traffic and weather, (2) statistical models that provide fine-grained incident likelihood forecasting, (3) resource allocation algorithms that dynamically optimize the spatial locations of responders and depots to improve response effectiveness, (4) algorithms that can provide early incident information by extracting insights from crowd-sourced data, and (5) algorithms for providing dispatch recommendations across the region, including support for coordination across multiple agencies in the community. 

\begin{figure}[t]
\begin{center}
\includegraphics[width=\columnwidth]{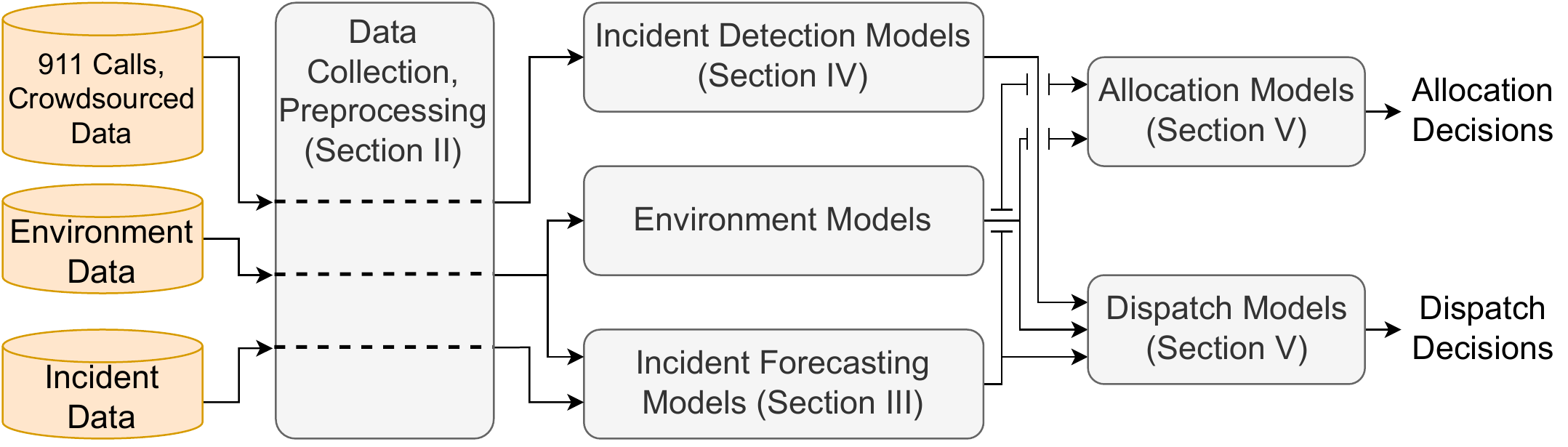}
\caption{ERM System Pipeline: historical data from different sources are used to design predictive models for incidents and the environment, which in turn are used to create allocation and response models. Events can be extracted using text and video data to expedite reporting and aid response.}
\label{fig:Sysmodel}
\end{center}
\vspace{-0.2in}
\end{figure}

Together, the goal of these components is to enable the agencies to (1) coordinate information and decisions between the many agencies involved with emergency response, (2) Collect, analyze and understand historical inefficiencies by creating data models using the diverse and high volume data generated from these events, (3) learn incident forecasting models that can generalize across large geographic areas, have high spatial-temporal resolution, and handle high data sparsity, (4) detect and report incidents automatically as quickly as possible, even from uncertain data-streams such as Waze and Twitter, and (5) dynamically adapt resource allocation and dispatch approaches, even if the environment in which the emergencies occur changes.

\input{tbl_challenges}

However, these goals are difficult to achieve, especially considering the following challenges. Principled emergency response systems need to account for the differing dynamics and variability of incidents. For example, the agencies have to respond to both daily incidents and large-scale disasters (e.g. natural hazards and man-made attacks) despite great variability in scales of impact for different categories of incidents. Disasters and security attacks can hinder operations and communications, which make centralized planning and cloud-based deployments infeasible.  In fact, the emergency response pipelines must be robust to any communication outage. Also, to improve allocation and response, agencies require incident forecasts over fine spatial and temporal resolutions. However, learning incident prediction models at high resolutions is extremely difficult due to data sparsity \cite{vazirizade2021learning}. 

Further, situational awareness during response requires information from heterogeneous data sources, e.g. weather, traffic, Twitter, Waze, 911 calls,  historical emergency data, and real-time crowdsourced data (e.g. near real-time road flooding). However, integrating varying forms of data into real-time incident detection models is highly non-trivial, especially when data exhibits noise and uncertainty ~\cite{senarath_emergency_2020,yasas2021,Purohit2018}. Lastly, it is critical that any learning procedure designed for emergency response be adaptive. That is, it should be able to provide recommendations even if the traffic patterns, housing density and the distribution of large events in the community change.

In this paper, we  describe some of these challenges and describe the approaches developed by our team to address them (see \cref{tab:challenges})\footnote{Our work has primarily focused on motor vehicle crashes, however, the approach we take is generalizable and apply to other incidents (except crime as it has different dynamics \cite{mukhopadhyay2016optimal}) that are responded to by our partners.}\footnote{For a comprehensive review of the state-of-the-art, please see our survey on incident prediction, resource allocation, and dispatch models~\cite{mukhopadhyay2020review}.}. \Cref {fig:Sysmodel} presents our overall ERM pipeline. In particular, 
section~\ref{sec:data} discusses the challenges involved in collecting and processing data for ERM systems. Section~\ref{sec:forecast} presents our framework for using this sparse incident data to create useful incident forecasting models. Section~\ref{sec:detection} discusses how to use crowdsourced data to detect incidents quickly. Section~\ref{sec:allocation} discusses the challenges in designing adaptive resource allocation and dispatch techniques that are robust to communication failures. Finally, section~\ref{sec:conclusion}  discusses open research questions that require further study.

%% file: tbl_challenges.tex
\begin{table*}[ht]
\centering
\caption{Summary of challenges when designing and deploying an emergency response management framework.  }
\vspace{-0.05in}
\label{tab:challenges}
\begin{tabular*}{\textwidth}{l@{\extracolsep{\fill}}lll}
\hline
\textbf{\textbf{Challenge}}                                     & \textbf{\textbf{Description}}                                                                                                                                                                                                                                                                                                                    & \textbf{\textbf{Contribution}}                                                                                                                                                                                                                                                                                                                                                \\ \midrule
\begin{tabular}[c]{@{}l@{}}Coordina- \\ tion\end{tabular}                                                & \begin{tabular}[c]{@{}l@{}}ERM requires coordination between multiple agencies \\ and decision makers, each with their own objectives. \\ Each decision maker often has access to only incomplete \\ information, and coordination must happen quickly while \\ a situation is unfolding.\end{tabular}                                           & \begin{tabular}[c]{@{}l@{}}ERM requires coordination across jurisdictions and agencies. We have\\ shown that response times to incidents decrease when resources dedicated \\ to a sub-region of a city ``help'' other regions in response~\cite{pettet_hierarchical_2021}. The strategy\\ for such coordinated response is an open challenge.\end{tabular}\\\hline
\begin{tabular}[c]{@{}l@{}}Data \\ Collection\end{tabular}      & \begin{tabular}[c]{@{}l@{}}It is difficult to collect, integrate, and preprocess the \\ eclectic data that forms the foundation for emergency \\ response systems. Much of the data has high volume and \\ velocity, and is from diverse sources. This large set of data \\ must then be narrowed down to a set of useful features.\end{tabular} & \begin{tabular}[c]{@{}l@{}}We have compiled and released WildfireDB, the first comprehensive \\ and open-source dataset that relates fires with relevant covariates~\cite{wildfiredb2021}.  \\ We have also designed data collection and feature engineering \\ pipelines for road accident data~\cite{vazirizade2021learning}.\end{tabular}                               \\\hline
\begin{tabular}[c]{@{}l@{}}Incident \\ Forecasting\end{tabular} & \begin{tabular}[c]{@{}l@{}}Incident occurrence is difficult to model due to incidents’ \\ inherent randomness and high sparsity. We have also \\ shown that Incident models are sensitive to spatial-\\ temporal resolution, which makes high-fidelity models \\ challenging to learn.\end{tabular}                                              & \begin{tabular}[c]{@{}l@{}}We have developed an incident prediction pipeline that combines \\ sparsity mitigation techniques with various statistical and learning \\ based forecasting models~\cite{vazirizade2021learning}. We \\ integrate this pipeline directly with decision making approaches to \\ evaluate the end-to-end performance of each model.\end{tabular} \\\hline
\begin{tabular}[c]{@{}l@{}}Incident \\ Detection\end{tabular}   & \begin{tabular}[c]{@{}l@{}}Fast incident detection is critical for timely response, but \\ traditional reporting methods have time delays. \\ Crowdsourced datastreams (e.g. Waze) provide an \\ opportunity for early identification, but are noisy and \\ uncertain.\end{tabular}                                                              & \begin{tabular}[c]{@{}l@{}}We have developed CROME, a novel practitioner-centered incident \\ detection approach using crowdsourced data~\cite{yasas2021}.\end{tabular}                                                                                                                                                                                                    \\\hline
\begin{tabular}[c]{@{}l@{}}Dynamic \\ Environm\\ents\end{tabular} & \begin{tabular}[c]{@{}l@{}}The environments in which ERM systems operate change \\ over both long and short time scales. ERM systems must \\ adapt to this nonstationarity.\end{tabular}                                                                                                                                                         & \begin{tabular}[c]{@{}l@{}}We have developed scalable online planning approaches that can \\ adapt to nonstationary environments, such as our hierarchical \\ ambulance allocation framework~\cite{pettet_hierarchical_2021}.\end{tabular}                                                                                                                                 \\\hline
\begin{tabular}[c]{@{}l@{}}Commun- \\ ications\end{tabular}                                            & \begin{tabular}[c]{@{}l@{}}Many emergency incidents cause failures in communication \\ networks. ERM systems must be robust to communication \\ loss to maintain service in such situations.\end{tabular}                                                                                                                                        & \begin{tabular}[c]{@{}l@{}}We have developed a partially decentralized approach to ambulance \\ allocation that allows ambulances to operate with limited \\ communication~\cite{pettet_algorithmic_2020}.\end{tabular}                                                                                                                                                    \\ \bottomrule

\end{tabular*}
\vspace{-0.2in}
\end{table*}

%% file: DataCollection.tex
\section{Data Collection and Integration}\label{sec:data}



\subsection{Challenges in data processing}
The collection of diverse, high-volume geospatial data can pose challenges for designing effective emergency response management systems.
Traditional desktop computers can be quickly overwhelmed by large volumes of geospatial data, necessitating the use of server computers with large amounts of main memory, a cluster of desktop computers, or cloud computing resources.
These resources can be financially prohibitive to smaller government organizations. Thus, precise programs are essential to minimize costs.
These programs will inevitably contain spatial operations or queries to a spatial database.
Spatial Operations and queries to spatial databases are computationally expensive, most notably spatial range, join, and k-nearest neighbor operations \cite{geospark}.
In an emergency response system, a variety of geospatial data formats (vector and raster) may arise, necessitating complex solutions: NoSQL data-stores or complex data structures, which can add additional processing time for the CPU \cite{li2016geospatial}.

To illustrate the problem, consider the context of wildfire response.
Data regarding fire occurrence and relevant features are often collected and stored by different agencies and sources. Importantly, such data is usually stored in different data models. For example, while fire occurrence data is usually stored in the vector form, information about vegetation, fuel, and topographic features is available in a raster model. These two data models use different storage mechanisms and computational methods that make it difficult to combine them. Also, collecting information about all relevant features through high spatial and temporal periods is challenging and mining such large-scale feature data is a massive computational bottleneck. To tackle such issues, we designed \textit{WildfireDB}, a collection of over 17 million data points that capture fire spread in continental United States in the last decade~\cite{wildfiredb2021}. Our data generation, to a large extent, is based on work on large-scale vector and raster analysis~\cite{singla2021raptor}, and uses a principled algorithmic approach to merge large-scale raster and vector data. \textit{WildfireDB} can be used to model the spread of wildfires and inform response strategies~\cite{diao2020uncertainty}. We manage incident data collection and joins in a similar manner.

\subsection{Data Sources}
Relevant and accurate data is essential for performant statistical models.
In our work in incident forecasting, we have relied on four primary types of data:
automobile incidents, roadway geometry, traffic, and weather data.
Past automobile accidents are relevant to the forecasting of future incidents because segments with a high recent incident rate are more likely to have incidents in the future \cite{mukhopadhyayAAMAS17, MukhopadhyayICCPS, mukhopadhyayAAMAS18}.
Spatial roadway features like lanes, speed limit, and road intersections often signify different driver behavior and, along with segment shape, are effective predictors of automobile incidents \cite{shankar1995effect, poch1996negative, chin2003applying}.
Weather data is valuable as precipitation, visibility, snowfall, and light levels can potentially alter driver behavior and is a valuable predictor of accident rates \cite{qi2007, songchitruksa2006assessing, mukhopadhyayAAMAS17}.
Traffic data is critical because it gives information about how drivers behave on roadways through speed and free-flow speed.
Traffic data also adds congestion, a measurement we have found to be one of the most predictive features.

%% file: IncidentForecasting.tex
\section{Incident Forecasting} \label{sec:forecast}


Incident forecasting is necessary to understand the future demand of emergency resources in a given region, which is critical for a proactive ERM system. 
Road accident prediction is well studied in the literature, and a variety of modeling approaches having been applied in the domain. These include statistical models such as hierarchical Poisson Models~\cite{Deublein2013, Quddus2008} and zero-inflated models~\cite{qin2004selecting, huang2010modeling}. In recent years, data mining models such as neural networks~\cite{Pande2006,zhu2018use,Bao2019} and support vector machines~\cite{doi:10.3141/2024-11,Yu2013} have also been explored. We refer interested readers to our survey on emergency response for a detailed analysis of prior work in forecasting road accidents (as well as allocation and dispatch techniques)~\cite{mukhopadhyay2020review}.

A shortcoming of prior accident forecasting approaches is their inability to deal with large data sparsity. Roadway accident data is extremely sparse -- we have observed more than $99\%$ sparsity on interstate highway segments in Tennessee (a state in the USA with a total area of over 100,000 sq. km.) when discretized into four hour time windows (more details can be found in paper~\cite{vazirizade2021learning}). Even zero-inflated models, the only class of statistical models that have been shown to work fairly well on sparse spatial-temporal data, fail to perform well when trained on these sparse roadway incidents. Prior approaches are therefore ill-equiped to make accurate short-term predictions, since as the temporal resolution increases, the sparsity increases as well. While long-term predictions can be useful to analyze policies (optimize road construction, for example), short-term predictions are desirable for allocating and dispatching responders.



\begin{figure}[t]
\centering
\begin{center}
\includegraphics[width=0.85\columnwidth]{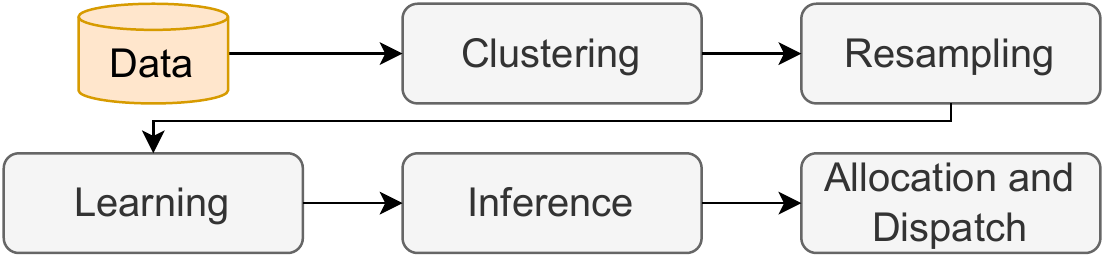}
\caption{
Overview of our incident prediction approach. We extract spatial temporal information from a variety of data sources, focus on heterogeneity not explicitly modeled in the feature space by identifying clusters, perform resampling to address sparsity, and learn models on incident occurrence. 
}
\label{fig:prediction_outline}
\end{center}
\vspace{-0.2in}
\end{figure}

To address these challenges, we have developed an incident prediction pipeline that integrates various sparsity mitigation and modeling techniques, as shown in figure \ref{fig:prediction_outline}. The pipeline consists of five major stages: (1) clustering, (2) resampling, (3) learning, (4) inference, and (5) integration with allocation and dispatch algorithms. First, we use clustering to identify spatial segments with similar patterns of incident occurrence. This aggregates the data for each cluster to avoid overfitting on individual segments while capturing heterogeneity that is not explicitly modeled in the feature space. 



Next, we perform synthetic under- and over-sampling to address sparsity by ``balancing'' the data. Naive synthetic sampling performs poorly, however, as the relative frequencies of incident occurrence vary significantly across clusters. We address this by performing resampling such that the ratio of accidents between each cluster is the same as in the original dataset. Specifically, we start with the cluster having the highest frequency of incident occurrence (cluster A, say) and perform synthetic resampling such that the number of positive data points (spatial segments in temporal windows that \textit{have} accidents) is the same as the number of negative data points (spatial segments in temporal windows that \textit{do not} have accidents). We then apply resampling to the other clusters such that the ratio of accident frequency for any given cluster compared to cluster A is the same as in the original dataset.


After clustering and resampling, we apply and compare various well-known spatial-temporal forecasting models to the processed dataset, including logistic regression, zero-inflated Poisson regression, random forests, and artificial neural networks. 
We use a suite of metrics to compare the performance of each model. Conversations with first responders revealed that \textit{ranking} roadway segments based on relative risk is useful for resource allocation. Therefore, we report the correlation of each model's marginal accident likelihood distribution over space with the real accident distribution in addition to standard statistical metrics like F1-scores and accuracy. Another important consideration is that these models are part of a pipeline that informs allocation decision making algorithms. Rather than only assess the models in isolation, we integrate them with an allocation decision agent and report each model's final impact on response times to simulated incidents.


We evaluated the forecasting pipeline on three years of road accident data from the interstate highway network of Tennessee and found it can significantly reduce incident response times compared to current approaches followed by first responders~\cite{vazirizade2021learning}. An important result of our work is the finding that standard statistical metrics like F1-score do not give an accurate representation of each model's performance when integrated with the full ERM pipeline: when used for resource allocation, the models which performed best in terms of these metrics ended up reducing response times less than models with `worse' scores. This result highlights the need to treat ERM pipelines holistically during both development and deployment.



%% file: IncidentDetection.tex
\section{Incident Detection}\label{sec:detection}

Emergency response is highly dependent on the time of incident reporting. However, the traditional approach of receiving incident reports (e.g., calling 911 in the USA) has time delays. Crowdsourcing platforms such as Waze provide an opportunity for early identification of incidents. However, detecting incidents from crowdsourced data streams is difficult due to the noise and uncertainty associated with such data. Further, simply optimizing over detection accuracy can compromise spatial-temporal localization of the inference, thereby making such approaches infeasible for real-world deployment.

We designed \textit{CROME} (Crowdsourced Multi-objective Event Detection), that quantifies the relationship between the performance metrics of incident classification (e.g., F1 score) and the requirements of model practitioners (e.g., 1 km. radius for incident detection)~\cite{yasas2021}. First, we show how crowdsourced reports, ground-truth historical data, and other relevant determinants such as traffic and weather can be used together in a Convolutional Neural Network (CNN) architecture for early detection of emergency incidents. However, naively maximizing the accuracy of detection can harm emergency response. For example, consider an approach that raises an alert every time a report is received on the crowdsourcing platform. While such an approach will correctly detect many (if not most) ground-truth incidents, it will lead to wastage of resources if a responder is dispatched on every alert. To tackle this challenge, we use a Pareto optimization-based approach to optimize the output of the CNN in tandem with practitioner-centric parameters to balance detection accuracy and spatial-temporal localization. We evaluated the efficacy of CROME using crowdsourced data from Waze and traffic accident reports from Nashville, TN, USA~\cite{yasas2021}. Our findings suggest that CROME can detect more than 40\% of the incidents earlier than traditional mechanisms.

%% file: ResourceAllocation.tex
\section{Resource Allocation and Dispatch} \label{sec:allocation}

Once we have an understanding of when and where incidents are likely to occur, the final stage of the ERM pipeline is using forecasting models to make \textit{allocation} and \textit{dispatch} decisions. Resource allocation (also referred to as the stationing problem~\cite{pettet_algorithmic_2020,mukhopadhyay2016optimal}) is the problem of spatially distributing resources such as ambulances in anticipation of stochastic incidents, while the dispatch problem is determining which resource to send to an incident once a call for service is received. The distinction between these problems is not always clear, as allocation policies create the implicit dispatching policy of sending the nearest available responder to an incident (greedy dispatch). In fact, conversations with emergency response agencies reveal that there are also legal constraints on dispatching models -- it is very difficult to judge the severity of an incident from a call for service~\cite{pettet_algorithmic_2020}, so it is imperative to follow this greedy strategy to ensure that if the incident is severe, a responder arrives as quickly as possible. Since dispatching policies are constrained by policy, we focus on solving the allocation problem.

A popular strategy used to plan under uncertainty is to learn a \textit{policy}, which is a general mapping from states of the environment to actions that should be taken. Reinforcement Learning and Approximate Dynamic Programming are examples of this approach, and have recently been applied to ambulance allocation and similar resource allocation problems such as ride-pooling~\cite{maxwell_ambulance_2009, nasrollahzadeh_real-time_2018, yu_integrated_2020, shah_neural_2020}. Unfortunately, these methods are slow to converge when applied to large-scale, practical problems. This certainly applies to ERM -- for example, a simple case of just $20$ responders and $30$ potential waiting locations has $\frac{|R|!}{(|R|-|V|)!} = \frac{30!}{10!} = {7.31e+25}$ possible assignments at each decision point. Since urban environments are nonstationary, a policy might be out of date by the time it is deployed. Policies can also have issues adapting to failures that were not considered during training.


An alternative to learning a policy is online planning: given a model of the environment, heuristic search approaches such as Monte-Carlo tree search (MCTS) can find promising actions for the current system state. The environmental models can be updated to reflect any detected changes, and the planner can incorporate these updates into the decision making process to adapt to a dynamic environment. Such an approach has been applied to real-time control problems in domains such as the smart grid~\cite{shuai_online_2021}, game playing~\cite{silver_mastering_2016}, and autonomous driving~\cite{hoel_combining_2020}. However, state-of-the-art MCTS approaches have difficulty converging in a reasonable amount of time for problems with large state-action spaces such as responder allocation. Another consideration is that many emergency incidents can cause failures in communication networks, so an allocation strategy should be able to operate with limited communication. Therefore, scalable online planning approaches that are robust to communication outages are needed for ERM.  


\begin{figure}[t]
    \centering
    \includegraphics[width=0.9\columnwidth]{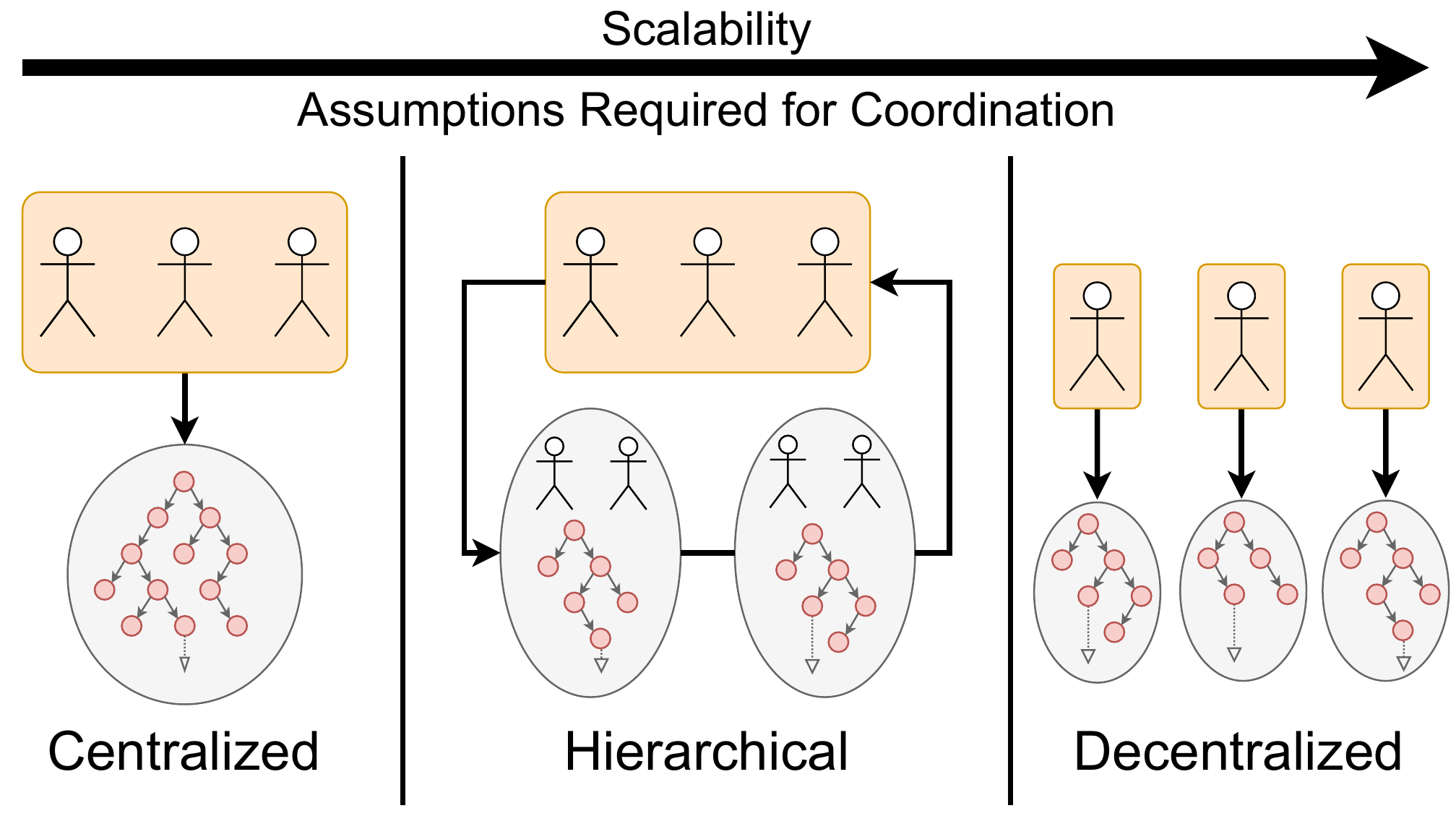}
    \caption{A spectrum of planning approaches for online ERM planning, each with inherent tradeoffs. A centralized approach uses a monolithic state representation. A decentralized approach segments the state-action space for each individual agent, and each agent performs their own decision making. A hierarchical approach segments the planning problem into sub-problems to improve scalability without agents estimating other agents' actions.}
    \label{fig:planning_spectrum}
     \vspace{-0.15in}
\end{figure}


We have developed two approaches to tackle the problem of scalable online resource allocation, as shown in figure~\ref{fig:planning_spectrum}. The first is decentralized planning, where each resource builds its own decision tree to independently determine its actions~\cite{pettet_algorithmic_2020}. By limiting the state-action space for each resource to only information relevant to the agent, the overall size of the state-action space is significantly decreased and the approach scales notably better than centralized approaches. A decentralized approach also addresses partial communication losses: since most mobile resources like ambulances are now equipped with compute units, they can perform their individual planning locally, only needing to communicate sporadically for coordination. Unfortunately, there is also a downside to decentralization -- during individual planning, resources must make assumptions regarding the behavior of other resources, which can lead to sub-optimal decisions when those assumptions do not hold. This tradeoff is worthwhile for problems like ERM that are difficult to solve tractably with centralized approaches. 



The second approach we developed is hierarchical planning, which leverages the spatial structure of the ERM environment to split a large resource allocation system into smaller sub-problems~\cite{pettet_hierarchical_2021}. This decomposition drastically decreases the overall complexity of the planning problem compared to a centralized approach, but preserves the dependencies between nearby resources unlike a decentralized approach. We created a principled framework for solving the sub-problems and tackling the interaction between them.
First, the overall spatial area is divided into regions using k-means clustering on historical data of incident occurrence. A high-level planner assigns responders to each region based on estimated waiting times using a queueing approximation. Then, a low-level planner solves the allocation problem within each sub-area in parallel using MCTS. Compared to a decentralized framework, a hierarchical approach is more reliant on communication channels within a region and is less scalable, but better models inter-resource dependencies. 

We applied our decentralized and hierarchical planning frameworks to the problem of dynamic ambulance allocation in the city of Nashville, TN. We found that the decentralized approach easily scaled to the full ERM system with reasonable computation times, and reduced mean incident response times compared to centralized planning~\cite{pettet_algorithmic_2020}. The hierarchical framework was also able to scale to the full system, and further improved average response times compared to the decentralized approach due to making fewer assumptions regarding the agents' interactions~\cite{pettet_hierarchical_2021}. We also found that both approaches are able to adapt to changes in the environment, such as large shifts in the incident distribution or ambulance failures.





%% file: Conclusion.tex
\section{Research Opportunities and Discussions}\label{sec:conclusion}

While tremendous progress has been made in designing emergency response management systems, the research area of building and deploying efficient and effective ERM in practice is still in its infancy. In this paper, we presented our recent efforts as the first step in developing key ERM components in collaboration with our community partners. There are many open research problems and real-world challenges in this exciting new area that need further studies. 

\textit{Uncertainty and guarantee}: Many current approaches have been designed to be robust to uncertainty and noise in the existing data. However, they barely deal with the uncertainty from other services/systems or human behavior in the real world, which could significantly affect the performance of incidents prediction and resource allocation. It is an open question of how to model the behaviors of a complex and integrated system considering this real-world uncertainty and, more importantly, providing run-time guarantees. 

\textit{Robustness}: Handling non-stationarity and environmental disruptions that will very likely reduce the effectiveness of the decision support algorithms are a key concern. In this context, it is clear that the online tree search algorithms are superior to a pure offline learning-based strategy that relies on reinforcement learning or a neural network approach. However, it is not yet clear if a combined approach that bootstraps the tree search can lead to improved outcomes. While preliminary experiments conducted by the team do indicate that such an approach can help with non-stationarity, further investigation is required.

\textit{Interaction with domain experts}: Despite the increasing intelligence in ERMs, there is still a high demand for inputs and interaction from the domain experts, e.g., emergency responders and city decision-makers. Two critical questions are (1) how to support experts to use our systems (e.g., providing requirements and domain knowledge), and then convert the information to machine-understandable language, and (2) how to help decision-makers interpret and trust the outcomes and decisions made by the ERM systems? 

\textit{Fairness and equity}: As data-driven approaches, machine learning models rely significantly on existing data. If unfairness exists in historical data, it is very challenging or impossible to learn fair models without exogenous intervention. Therefore, measuring and validating group fairness and addressing equity when designing ERM pipelines is still an important yet challenging future direction. 

\textit{Practical Constraints}: Another issue that is an opportunity for further research is the integration of practical constraints into the decision support procedures. Such dynamic considerations include reduction in the number of ambulances available, constraints on the maximum mileage that can be put on a vehicle, road closures and reduced manpower. Recent experiments by Ferber et.al. in \cite{ferber2020mipaal} have shown that including the decision outcomes in the prediction pipeline can lead to improved forecasting models. Such considerations may also be better in integrating various practical constraints that can be encountered. 


\section*{Acknowledgements}

This work is sponsored by the National
Science Foundation under award numbers CNS1640624, CNS1818901, IIS1814958, IIS-1815459 and a grant from the Tennessee Department of Transportation.